\def\year{2021}\relax
\documentclass[letterpaper]{article} 
\usepackage{aaai21}  
\usepackage{times}  
\usepackage{helvet} 
\usepackage{courier}  
\usepackage[hyphens]{url}  
\usepackage{graphicx} 
\usepackage{natbib}  
\usepackage{caption} 
\usepackage{pgfplots}
\definecolor{oxfordblue}{HTML}{2f2adb}
\definecolor{darkgreen}{HTML}{078921}
\usepackage{color, colortbl}
\definecolor{Gray}{gray}{0.9}

\title{Combining Machine Learning and Human Experts to Predict Match Outcomes in Football: A Baseline Model\\
\small Challenging Problems Track}

\author {

        Ryan Beal, 
        Stuart E. Middleton, 
        Timothy J. Norman, 
        Sarvapali D. Ramchurn \\
}
\affiliations {
    School of Electronics and Computer Science, University of Southampton,\\
Southampton, SO17 1BJ, United Kingdom \\
     \{ryan.beal,sem03,t.j.norman,sdr1\}@soton.ac.uk
}

\begin{document}

\maketitle

\begin{abstract}
In this paper, we present a new application-focused benchmark dataset
and results from a set of baseline Natural Language Processing and Machine Learning
models for prediction of match outcomes for games of football (soccer). By doing so we
give a baseline for the prediction accuracy that can be achieved exploiting both statistical
match data and contextual articles from human sports journalists. Our dataset is focuses
on a representative time-period over 6 seasons of the English Premier League, and includes
newspaper match previews from The Guardian. The models presented in this paper
achieve an accuracy of 63.18\% showing a 6.9\% boost on the traditional statistical methods.
\end{abstract}

\section{Introduction}\label{sec:Intro}

Real-world events such as sports games or elections involve competing teams, each with capabilities and tactics, aiming to win (e.g., seats during an election, or scoring more goals in a football match). The performance of such teams is typically not only dependent on the teams' abilities but also on the environment within which they operate. For example, a political party may have the best orators and policies but their opponents may be better at getting votes in key areas. Similarly, a top football team may be playing the worst team in a league but the fact that the latter may be facing relegation (to a lower league) may provide them with extra motivation to win the game. Given this, in many cases, the performance of such teams may not be easily predictable.  


Traditional AI and machine learning techniques to predict the outcome of real-world events tend to focus on the use of statistical machine learning using historical data about the individual teams \cite{silver2012signal,campbell1988stock,dixon1997modelling,matthews2012competing}.  However, as per the examples above, historical performance may not be useful when team performance may be dependent on dynamic factors such as human performance (morale, injuries, strategies) or environmental variables (weather, competition context, public mood). In turn, humans can be better judges than algorithms when faced with previously unseen situations. Journalists, online communities, and experienced analysts may be better at evaluating human and environmental elements to forecast an outcome. For example, one approach of looking at more than just statistics in sports have been through sentiment analysis on social media platforms. Schumaker, Jarmoszko and  Labedz (\citeyear{schumaker2016predicting}) use this approach to predict English Premier League (EPL) results and achieve an accuracy of 50\% and \cite{sinha2013predicting} show use of similar analysis being performed for American Football results in the National Football League (NFL) predicting the winner 63.8\% of the time. However, these approaches focus on opinion aggregation rather than trying to extract the potential indicators of performance for individual human teams from human experts.

Against this background, we set new baselines for results when predicting real-world sporting events involving humans based on the combination of Natural Language Processing (NLP) and statistical machine learning techniques. In more detail, we focus specifically on football games in the EPL using match previews from the media alongside statistical machine learning (ML) techniques. The prediction of football match outcomes is a challenging computational problem due to the range of parameters that can influence match results. To date, probabilistic methods devised since the seminal work of Maher (\citeyear{maher1982modelling}) have generated fairly limited results and appear to have reached a glass ceiling in terms of accuracy. By using media previews we can improve on the accuracy of current approaches for match outcome prediction. By so doing, we show that by incorporating human factors into our model, rather than just basic performance statistics, we can improve accuracy (e.g., mood, rivalries and other external factors). Thus, the contributions of this paper are as follows:

\begin{enumerate}
\item New dataset for testing NLP/ML algorithms for sports match outcome prediction for football (soccer). Our dataset includes a previously unexplored feature set in terms of football match outcome predictions, including human knowledge that is overlooked in traditional statistics. The dataset includes match data and previews for 1770 games football games over 6 seasons.
\item Set of baseline models using a novel mixture of OpenIE, text vectorisation and supervised ML methods for predicting the outcome of games of football using human opinions from domain-experts in the media.
\item Set of benchmark results for our baseline algorithms predicting the outcomes of 1770 games and additional results, including results for more traditional statistical approaches and baseline predictions from bookmakers’ odds (i.e. human predictions).
\end{enumerate}

In the next section we discuss the match outcome prediction problem for football and the new feature set we explore.



\section{The Problem}\label{sec:problem}

Team sports games presents to us a challenging problem of predicting what the outcome of the game will be based on the previous performance of the teams and the relative abilities of the players in the teams. On paper, this should be quite simple, the team with the best players should win. In the real-world however, this is not always the case and achieving high accuracies can be a serious challenge. Bookmakers run their businesses based on this challenge and use sophisticated pricing models that assign “odds” to an outcome (which reflect the likelihood) to maximise their chances of making a profit. As discussed in \cite{beal2019artificial}, the accuracy of bookmakers is at around 67\% for American football, 74\% for basketball, 64\% in cricket, 61\% in baseball and at just 54\% for football. 

Football (soccer) therefore presents to us the most interesting prediction problem in team sports. Due to the low scoring nature of the game we see many more surprising results, where a lower quality team can win against a team made up from world class players. In comparison to the other sports we mentioned, there are far more draws in football as well, meaning that there are three possible outcomes when we are making predictions. There are also a range of uncertainties that can influence match results including the team configurations, the health of players, the location of the match (home or away), the weather, and team strategies. Thus making the prediction of football match outcomes a very complex computational problem. 

Beal, Ramchurn and Norman (\citeyear{beal2019artificial}) discuss the many different applications in machine learning and statistics that have been applied to the challenge of predicting sports outcomes. They find that these applications with standard statistical numerical inputs have reached a ``glass ceiling" in terms of prediction accuracy. They show \cite{baboota2019predictive} and \cite{dixon1997modelling} are two of the most accurate models to date, showing accuracies on average at around 56.7\% and 59.1\% respectively. Therefore, in this paper we set a new baseline by assessing the use of a new feature set. These features will be taken from the match preview reports written by human expert journalists which aim to incorporate some intangible variables that cannot be factored into traditional statistics.

\section{Modelling Human Opinion}\label{sec:Model}

In this section we discuss how we model the problem of extracting information from media articles written by expert football writers and how we solve this model.

\subsection{The Model}

Each game of football that we aim to predict involves two teams competing to win a game (home team $\alpha$ and away team $\beta$) and there are a set of 3 possible outcomes - home win, draw, away win $\mathcal{O} = \{homewin, draw, awaywin\}$. Each game has a set of pre-game preview articles written about it $T=\{t_1, t_2, ..., t_j\}$ where $j \ge 1$. In this paper, we focus on just one of the sources for the pre-game preview articles (from the Guardian website), but this could be expanded to include as many of the articles in $T$ as possible. Another example of this type of analysis could be applied to elections, in an election, the possible outcomes would be the different political parties that could win.

Each game-preview article text $t \in T$ is built up from multiple sentences denoted by $t=\{s_1, s_2,..., s_k\}$ where $k$ is the number of sentences in the text article. We transform each sentence $s$ into a form that we can use as a feature to give $f(s)$, where $f$ represents the transformation function to output a numeric vector representation of the sentence.

Each sentence may be related either team $\alpha$ or $\beta$ competing in the game. We must also consider that the sentence is related to neither of the teams or both of the teams. We therefore calculate the probability that each sentence relates to a team competing in the game and then allocate that sentence to the most likely team (e.g., $p(f(s)|\alpha)$) . If there is a high level of uncertainty in the allocation then the allocation is set to no team. An allocation is defined as a pair of the sentence vector and its allocated team (home team, away team, no team), $a = (f(s), \{\alpha \lor \beta\})$ where $a \in \mathcal{A}$ and $\mathcal{A}$ is the set of allocations $|\mathcal{A}|=k$. 

With this set of allocations, we calculate the final features ($X$) for each of the teams to allow us to make a prediction on the outcome of the game $y$ where $y \in \mathcal{O}$. We can use these text-based features in $X$ to train a machine learning model to predict $y$. The features are created by an addition of all the sentence vectors that are allocated to one of the teams $\alpha$ or $\beta$ which gives the final vector for each team $V(\alpha)$ and $V(\beta)$. For the problem of predicting football we must also include a home team advantage weighting ($\mu$) to our feature set as discussed in \cite{clarke1995home}. We now define our feature set $X$ for each game as $X=[\mu \cdot V(\alpha), V(\beta)]$ and the target $y$ is the actual outcome which we are aiming to predict using a Random Forest method $\phi$ so that $\phi(X) = y$.  

\subsection{Extracting Human Opinion}

\begin{figure}[h!]
    \centering
    \includegraphics[scale=0.85]{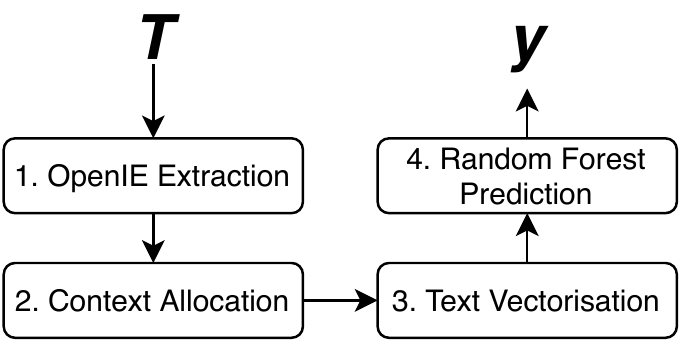}
    \caption{Model Process Diagram.}
    \label{fig:process}
\end{figure}

To extract the human opinion based features and solve our model we start with some set of texts $T$ from a given source and we are aiming to produce a prediction of the outcome of a game of football, $y$ that the texts relate to. Specifically, we apply the steps which are outlined in Figure \ref{fig:process}. Here, we discuss each stage and the methods that we use.

\begin{enumerate}
    \item \textbf{OpenIE Extraction:} Relation tuples are extracted in the form \{argument, relation, argument\} for each sentence in the articles text, for example \{Machester United, ex-manager, Mourinho\}.
    \item \textbf{Allocation of Text Context:} Each sentence is allocated to a team ($a = (f(s), \{\alpha \lor \beta \})$). For example, in football, each sentence must be allocated to one of the teams that are playing in the match that the article is discussing. 
    \item \textbf{Text Vectorisation:} We convert the sentences into vectors using a Count Vectorizer technique so we have a numerical representation of the words in a sentence. These give $f(s)$ and the final features are computed to form $X$.
    \item \textbf{Prediction:} Once we have formed our feature set ($X$) for each game that we are aiming to predict, we train a Random Forest model ($\phi$) using historic data (with outcomes) and the features $X$. This is used to make the final predictions of events based on the original articles. The outcome that we aim to predict ($y$) corresponds to one of the possible outcomes in $\mathcal{O}$ (e.g., for football predictions $y = {homewin, draw, awaywin}$).
\end{enumerate}

This process is generalizable and could be used across a number of domains in sport and beyond to predict real-world team events (e.g., elections). In the next section we discuss the text dataset that we extract.

\section{Text Dataset from the Media}\label{sec:data}

As well as the model outlined in this paper, we provide a novel dataset for researchers to use and implement their own models for the challenge of predicting sports events using human experts from the media. This dataset can be found at \textit{https://github.com/RyanBeal7/GuardianPreviewData} with all data sourced from the Guardians English Premier League match previews over the past 6 seasons.\footnote{Guardian articles are open source and can be found here: https://www.theguardian.com/football/series/match-previews.} The dataset does not contain an exhaustive list of all games during the seasons that we focus on, although we are able to source data from 1770 games across 6 seasons from 2013/14 to 2018/19. This paper shows the first analysis of a dataset of this type, combining text and statistics to predict football matches. In this domain, 1770 games is a large dataset for predictions of football games, other examples of papers for this problem usually only tests on 1 or 2 seasons of data (380 games per EPL season).  An example snippet from a match preview regarding a game between Southampton and Tottenham in 2019 is as follows:

\small \textit{``Which Tottenham team will show up at St Mary’s? Tottenham competed a clinical Champions League victory over Dortmund – but they are winless in their last three league games. Pochettino begins his touchline ban and will be without Kieran Trippier, although Dele Alli and Harry Winks could feature."}

\normalsize
This type of dataset can be used to  analyse how the wording and sentiment regarding teams in similar reports (over the 1770 games) correlates to the match outcome. We expect that human related factors (e.g., team rivalries and behind the scene changes in staff/players) will be brought through in our predictions and improve on the traditional statistical approaches. In the next section, we use this dataset to set baselines for this type of predictions. 



\section{Benchmark Performance}\label{sec:baseline}

This section outlines the experiments performed to benchmark the performance for the challenges discussed in this paper. We asses the outcome accuracy, ability to identify longshots/draws and the performance across a season, these allow us to show how the new text based features improve on standard approaches. Experiments are run using historic data taken from the Guardian match previews and historical bookmaker odds were taken from OddsPortal.\footnote{https://www.oddsportal.com/results/soccer.} All experiments are run using match previews written before the game took place and with pre-match bookmaker odds to ensure that each test is fair and there is no contamination of data in our experiments. 


\subsection{Experiment 1: Accuracy of NLP Outcome Prediction}
Using the methods that we have outlined in this paper we test and compare the accuracy, precision and recall against Dixon \& Coles (\citeyear{dixon1997modelling}) model and the bookmakers. Dixon \& Coles is still seen as one of the leading examples of statistical modelling for football and has not been significantly improved using machine learning techniques, this is discussed in \cite{beal2019artificial}. Therefore, we aim to improve on the Dixon \& Coles model, as well as the bookmakers accuracy.   

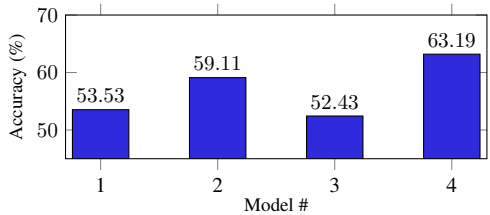
\begin{figure}[h!]
\centering
\begin{tikzpicture}[thick,scale=1, every node/.style={scale=0.8}]
    \centering
    \begin{axis}[
            ybar,
            bar width=0.75cm,
            symbolic x coords={1,2,3,4,5,6},
            xtick=data,
            ylabel={\small Accuracy (\%)},
            xlabel={\small Model \#},
            width=\columnwidth-35,
            height=\columnwidth-140,
            nodes near coords,
            nodes near coords align={vertical},
            ymin=45,ymax=70,
            y label style={at={(axis description cs:0.1,.5)},anchor=south},
        ]
        \addplot[fill=oxfordblue] table[x=interval,y=diff]{\mydata};
    \end{axis}
\end{tikzpicture}
\caption{Comparison of Model Accuracies.}
\label{fig:comp}
\end{figure}

The results from this test are shown in Figure \ref{fig:comp} and Table \ref{tab:rc}. The test was run taking an average of the performance across 3 seasons (2016-2019), using a training set of all games prior to that season and a test set of 300 games in the season. It is worth noting that the class distribution of EPL games from across 25 seasons is 46.2\% home wins, draws 27.52\% and away wins 26.32\%.

\noindent
\begin{itemize}
    \item \textbf{Model 1 (Text Vectors):} The process shown in Figure \ref{fig:process} with features formed using text vectorisation methods.
      \item \textbf{Model 2 (Dixon and Coles):} The model described by Dixon and Coles \cite{dixon1997modelling}.
       \item \textbf{Model 3 (Bookmakers):} Uses the pre-match bookmakers favourite as the predicted winner of the game.
        \item \textbf{Model 4 (Text Vector Combination):} Uses features calculated from text vectorisation, outputs from Dixon and Coles and the pre-match bookmakers odds. The features are the probability of a home win, away win and draw from each of the three models. Therefore, we have 9 features per game and apply a Random Forest classifier.
\end{itemize}

\pgfplotstableread[row sep=\\,col sep=&]{
    interval & diff \\
    1     & 53.53   \\
    2    & 59.11 \\
    3   & 52.43  \\
    4   & 63.19  \\
    }\mydata

\begin{table}[h!]
\caption{Precision/Recall Results.}
\centering
\begin{tabular}{|c|c|c|c|}
\hline
\rowcolor{Gray}
Model \# & Precision & Recall & F1 Score \\ \hline
1        &     \textbf{0.649}       &     0.413   &  0.505   \\ \hline
2        &      0.503           &    0.491   &  0.496     \\ \hline
3        &      0.451           &    0.452    &  0.451    \\ \hline
4        &     0.612            &    \textbf{0.563} &   \textbf{0.586}     \\ \hline
\end{tabular}
\label{tab:rc}
\end{table}

These results show that using NLP methods, on their own, do not produce remarkable results. Model 1 (using just text vector methods) produced an accuracy of 53.5\% which is not able to better the Dixon and Coles predictions. However, we found that when we use an ensemble learning approach and use the prediction probabilities output from Model 1 with the bookmakers probabilities and Dixon and Coles probabilities we can improve these methods. Model 4 (using all the models probabilities) achieves an accuracy of 63.2\% which is a 10.8\% increase on the bookmakers accuracy and 4.1\% more than Dixon and Coles (6.9\% boost). We show that Model 4 has the highest F1 score, which is 0.113 and 0.124 more than Dixon and Coles the bookmakers respectively. 

We also test how Model 4 performs without the use of the text vector features. This highlights that it is the new features which cause the boost in accuracy. We find that without the text features, the F1 score is 10\% less and the accuracy 7\% less than Model 4, and therefore show that increased performance is due to the text vector features. We also show that our model can outperform the sentiment analysis approach in \cite{schumaker2016predicting} by 13\%.

\subsection{Experiment 2: Longshots and Draws}

The traditional statistical models and bookmakers' approach to predicting football match outcomes are typically poor at predicting draws and longshot results. A longshot result is when the winning team has a bookmaker probability of less than 20\%. Therefore, we test the ability of our new approach to predict these events. To do this we split all 1770 games into training and test sets (random 80/20 split). In the test-set there are 75 draws and 47 longshot outcomes. In terms of draws, models 1,2 and 3 do not predict any of the draws. Whereas, Model 4 predicts 26.5\%. For the longshot results, model 1 predicts 38.9\%, models 2 and 3 do not predict any of the longshots and Model 4 predicts 22.2\%.

These results show that by using our model we can more accurately identify when draws and longshot results are likely to happen, this is especially the case where the text vectors model identifies more longshots (38.9\%) by taking into account the human input. This may be because there are some more subjective factors and knowledge that has an effect on games which cannot be considered in more statistical approaches. Some examples of this may be due to text articles that discuss the possible line-up of the team and if a manager may rotate or if a team signs a new manager/player.


\subsection{Experiment 3: Season Performance}

In this experiment, we assess how our ensemble learning model (Model 4) performs over an entire EPL season in comparison to Dixon and Coles and the bookmakers. To do this, we train our model using all data from the seasons before the 18/19 season (all articles and statistics) and then run our model for each game-week to show how many matches would be predicted correctly across the season. This is shown below in Figure \ref{fig:weeks} where we show the accumulation of correct results across the season.

\begin{figure}[h!]
\centering
\begin{tikzpicture}[thick,scale=1, every node/.style={scale=0.8}]
	\begin{axis} [
		xlabel=Gameweek,
		ymin=0,ymax=190,
		xmin=0, xmax=38,
		width=\columnwidth-20,
        height=\columnwidth-100,
        legend pos=north west,
        y label style={at={(axis description cs:0.075,.5)},anchor=south},
		ylabel= \# Correct Predictions]
		
	\addplot[color=red,line width=0.25mm, solid] coordinates{
		(0,0) (1,8) (2,15) (3,22) (4,27) (5,35) (6,42) (7,50) (8,54) (9,62) (10,68) (11,73) (12,78) (13,82) (14,91) (15,91) (16,97) (17,105) (18,109) (19,109) (20,115) (21,115) (22,119) (23,127) (24,127) (25,134) (26,140) (27,143) (28,143) (29,144) (30,150) (31,153) (32,160) (33,162) (34,169) (35,174) (36,178) (37,182) (38,186) 
	};
	\addplot[color=darkgreen,line width=0.25mm, dotted] coordinates {
		(0,0) (1,6) (2,10) (3,14) (4,17) (5,23) (6,29) (7,38) (8,42) (9,49) (10,53) (11,59) (12,65) (13,70) (14,76) (15,76) (16,80) (17,84) (18,87) (19,87) (20,91) (21,91) (22,96) (23,104) (24,104) (25,110) (26,115) (27,117) (28,117) (29,118) (30,123) (31,125) (32,132) (33,133) (34,138) (35,142) (36,144) (37,147) (38,150) 
	};
	\addplot[color=blue,line width=0.25mm, dashed] coordinates {
		(0,0) (1,8) (2,15) (3,22) (4,26) (5,33) (6,40) (7,49) (8,53) (9,60) (10,64) (11,70) (12,77) (13,82) (14,90) (15,90) (16,95) (17,103) (18,108) (19,108) (20,113) (21,113) (22,117) (23,126) (24,126) (25,130) (26,134) (27,137) (28,137) (29,138) (30,143) (31,145) (32,152) (33,154) (34,161) (35,165) (36,167) (37,172) (38,177) 
	};
	\addlegendentry{\small Ensemble}
	\addlegendentry{\small Bookmakers}
	\addlegendentry{\small Dixon \& Coles}
	\end{axis}
\end{tikzpicture}
\caption{2018/19 EPL Week by Week Analysis}
\label{fig:weeks}
\end{figure}
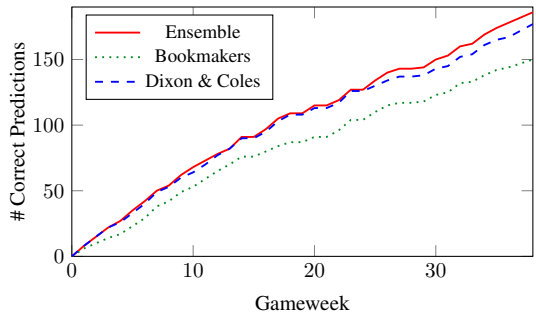

This shows that ensemble learning approach can continually perform well and predict correct results across a given EPL season when set up in a real-world scenario. We also find that there is a 2.23\% increase in accuracy between week 1 and week 38 for model 4, showing that as the season progresses our model improves. It also shows how the articles can better model the different scenarios that teams may be in later on the season which numbers do not represent as well. For example, if a team has played poorly all season but is now in a relegation battle they may have more to play for than a mid-table team with no chances of winning the league or relegation. The same goes when teams are fighting to win the league or qualify for European competitions. Injuries and rotation also play a big part later in the season as teams who have progressed into the later rounds of the FA Cup and European competitions have many more games. This shows the key contribution of the media preview analysis and how by taking into the human factors that are written about by the domain-expert journalists we are able to better predict the outcome of football matches.

\section{Conclusion}\label{sec:conclusion}

This paper has presented a novel application-focused dataset and has set new baselines of 63.19\% accuracy for predicting games of English Premier League football across a three season period using a novel dataset which we provide as part of this paper. We showed that the application of combining human opinion and machine learning to make predictions can boost the accuracy of traditional methods and those using sentiment analysis on social media. We show that we boost these methods by 6.9\% in terms of outcome accuracy and that the model accuracy increases as the season progresses and human factors/emotions begin to play a bigger part in the game.

\clearpage

\section*{Acknowledgements}

We would like to thank the reviewers for their comments. This research is supported by the AXA Research Fund and the EPSRC NPIF doctoral training grant number EP/S515590/1.

\bibliographystyle{aaai}
\bibliography{refs.bib}

\end{document}